\def\year{2021}\relax
\documentclass[letterpaper]{article} % DO NOT CHANGE THIS
\usepackage{aaai21}  % DO NOT CHANGE THIS
\usepackage{times}  % DO NOT CHANGE THIS
\usepackage{helvet} % DO NOT CHANGE THIS
\usepackage{courier}  % DO NOT CHANGE THIS
\usepackage[hyphens]{url}  % DO NOT CHANGE THIS
\usepackage{graphicx} % DO NOT CHANGE THIS
\usepackage{natbib}  % DO NOT CHANGE THIS AND DO NOT ADD ANY OPTIONS TO IT
\usepackage{caption} % DO NOT CHANGE THIS AND DO NOT ADD ANY OPTIONS TO IT
\usepackage{subcaption}

\usepackage{algorithm}  
\usepackage{algorithmicx} 
\usepackage{algpseudocode}

\usepackage{multirow}
\usepackage{booktabs}
\usepackage{colortbl} 
\usepackage{arydshln} 

\usepackage{amsmath}
\usepackage{amssymb}
\usepackage{verbatim}
\usepackage{siunitx}
\usepackage{bm}

\newcommand{\myparagraph}[1]{\noindent\textbf{#1}\hspace{1.8ex}}

\usepackage{appendix}
\title{Collaborative Group Learning}
\author {
        Shaoxiong Feng,\textsuperscript{\rm 1}
        Hongshen Chen,\textsuperscript{\rm 2}
        Xuancheng Ren,\textsuperscript{\rm 3}
        Zhuoye Ding,\textsuperscript{\rm 2}
        Kan Li,\textsuperscript{\rm 1}
        Xu Sun\textsuperscript{\rm 3,4} \\
}
\affiliations {
    \textsuperscript{\rm 1}School of Computer Science \& Technology, Beijing Institute of Technology
    \textsuperscript{\rm 2}JD.com \\
    \textsuperscript{\rm 3}MOE Key Laboratory of Computational Linguistics, School of EECS, Peking University \\
    \textsuperscript{\rm 4}Center for Data Science, Peking University \\
    \{shaoxiongfeng, likan\}@bit.edu.cn, \{renxc, xusun\}@pku.edu.cn \\
    ac@chenhongshen.com, dingzhuoye@jd.com
}

\begin{document}
\maketitle

\begin{abstract}
Collaborative learning has successfully applied knowledge transfer to guide a pool of small student networks towards robust local minima. However, previous approaches typically struggle with drastically aggravated student homogenization when the number of students rises. In this paper, we propose Collaborative Group Learning, an efficient framework that aims to diversify the feature representation and conduct an effective regularization. Intuitively, similar to the human group study mechanism, we induce students to learn and exchange different parts of course knowledge as collaborative groups. First, each student is established by randomly routing on a modular neural network, which facilitates flexible knowledge communication between students due to random levels of representation sharing and branching. Second, to resist the student homogenization, students first compose diverse feature sets by exploiting the inductive bias from sub-sets of training data, and then aggregate and distill different complementary knowledge by imitating a random sub-group of students at each time step. Overall, the above mechanisms are beneficial for maximizing the student population to further improve the model generalization without sacrificing computational efficiency. Empirical evaluations on both image and text tasks indicate that our method significantly outperforms various state-of-the-art collaborative approaches whilst enhancing computational efficiency.
\end{abstract}
\section{Introduction}
Deep neural network has achieved impressive performance in various fields. Combining multiple individual networks, an ensemble model gains better predictive performance than a single network. One important reason is that an ensemble model usually aggregates a robust local minimum rather than a sharp local minimum that a single model may be stuck in. To alleviate the prohibitive computational cost of those high-capacity ensemble networks, Knowledge Distillation (KD) method is therefore proposed to achieve more compact yet accurate models by transferring knowledge~\cite{DBLP:conf/nips/BaC14,DBLP:journals/corr/RomeroBKCGB14,DBLP:journals/corr/HintonVD15,han2015deep}. KD comprises two pipelined learning stages, a pre-training stage and a knowledge transfer stage. Recently, attempts on group-based online knowledge distillation, also known as collaborative learning, explore less costly and unified models to eliminate the necessity of pre-training a large teacher model \cite{DBLP:conf/cvpr/ZhangXHL18,DBLP:conf/iclr/AnilPPODH18}, where a group of students simultaneously discovers knowledge from the ground-truth labels and distills group-level knowledge (multi-view feature representation) from each other. 
 
Collaborative learning shares the benefits of finding a more robust local minimum than a single model learning while accelerating the model learning efficiency compared with conventional KD. In terms of the implementation of student networks in collaborative learning, DML \cite{DBLP:conf/cvpr/ZhangXHL18} uses a pool of network-based students, where each student is an individual network and they asynchronously collaborate, whereas CL-ILR \cite{DBLP:conf/nips/SongC18} proposed branch-based collaborative learning that all the student networks share the bottom layers while dividing into branches in the upper layers. Benefiting from representation sharing (an extreme form of hint training \cite{DBLP:journals/corr/RomeroBKCGB14}), CL-ILR not only is more compact and efficient but also shows better generalization performance. As observed in \cite{DBLP:conf/cvpr/ZhangXHL18,DBLP:conf/nips/SongC18,DBLP:conf/nips/LanZG18,DBLP:conf/aaai/ChenMWF020}, the model performance continually improves along with the increasing number of students.
 
However, the students in collaborative learning tend to homogenize, damage the generalization ability, and degrade to the original individual network. Although the students are randomly initialized, they learn from the same entire training set and are prone to converging to similar feature representations \cite{DBLP:journals/corr/LiYCLH15,DBLP:conf/nips/MorcosRB18}. Moreover, each student distills knowledge from all other students, which further aggravates the homogenization problem due to ignoring the diversity of students' group-level knowledge \cite{DBLP:journals/cim/Schwenker13,DBLP:conf/nips/LanZG18,DBLP:conf/aaai/ChenMWF020}. Another insurmountable obstacle for collaborative learning is that the computational cost boosts greatly as well when more students join in collaborative learning.

To overcome these challenges, in this paper, we propose a collaborative group learning framework that improves and maintains the diversity of feature representation to conduct an effective regularization. Intuitively, under the spirit of knowledge distillation by learning as collaborative groups, we divide the whole course into multiple segments and assign students into several non-isolated sub-groups. Each student learns one piece of course and grasps the whole course through efficient knowledge distillation. Specifically, we \textit{first} introduce a conceptually novel method, called \textbf{random routing}, to build student networks, where each student is regarded as a group of network modules, and the connections between network modules are established as randomly routing a modular neural network. After randomly routing, we break the limitation of sharing representation only at the bottom layers and extend its range to any layer. Modules are shared by different involved students, which facilitates knowledge sharing and distillation between students. \textit{Second}, to tackle the student homogenization problem aggravating with the increasing number of students, \textbf{sub-set data learning} is proposed for each student to learn different parts of the training set. It increases the model diversity by introducing the inductive bias of the data subset into the student training. Moreover, to compensate for the knowledge (training data) loss of individual students while maintaining the student diversity, we further propose \textbf{sub-group imitation}, where a sub-group of students is randomly selected and assigned to aggregate group-level knowledge in each iteration, rather than aggregating knowledge from all other students as in previous approaches. It allows a student to internalize dynamic and evolving group-level knowledge while adjusting the sub-group size for adapting to various computational environments. In addition to be collaboratively devoted to the student homogenization problem so as to regularize the feature learning effectively, the above three mechanisms also enhance computational efficiency (e.g., reduce the number of parameters and the number of forward and backward propagation), which means our framework can maximize the student population to further improve the model generalization under restricted computational resources.
 
In summary, our contributions are as follows: 1) Collaborative group learning strikes a better balance between diversifying feature representation and enhancing knowledge transfer to induce students toward robust local minima. 2) Random routing builds students by randomly connecting the module path, which enables random levels of representation sharing and branching. 3) To overcome the student homogenization problem, sub-set data learning first draws on the inductive bias from the data subset to improve the model diversity. Then, sub-group imitation further transfers supplementary knowledge from a random and dynamic sub-group of students, maintaining the diversity of students. 4) Besides, the proposed three mechanisms are computationally efficient, allowing more students to join in collaboration and further boost the model generalization. We conducted detailed analyses to verify the advantages of our framework on generalization, computational cost, and scalability.
\section{Method}
Compared with previous approaches, collaborative group learning has the superiority in generalization performance and computational efficiency. In this section, we first elaborate how to build students using \textbf{random routing}, then introduce \textbf{sub-set data learning} and \textbf{sub-group imitation} to discover and transfer diverse knowledge effectively, and finally present the training objective.

\subsection{Random Routing for Student Network}
\begin{figure}[t]
    \centering
    \includegraphics[width=0.85\linewidth]{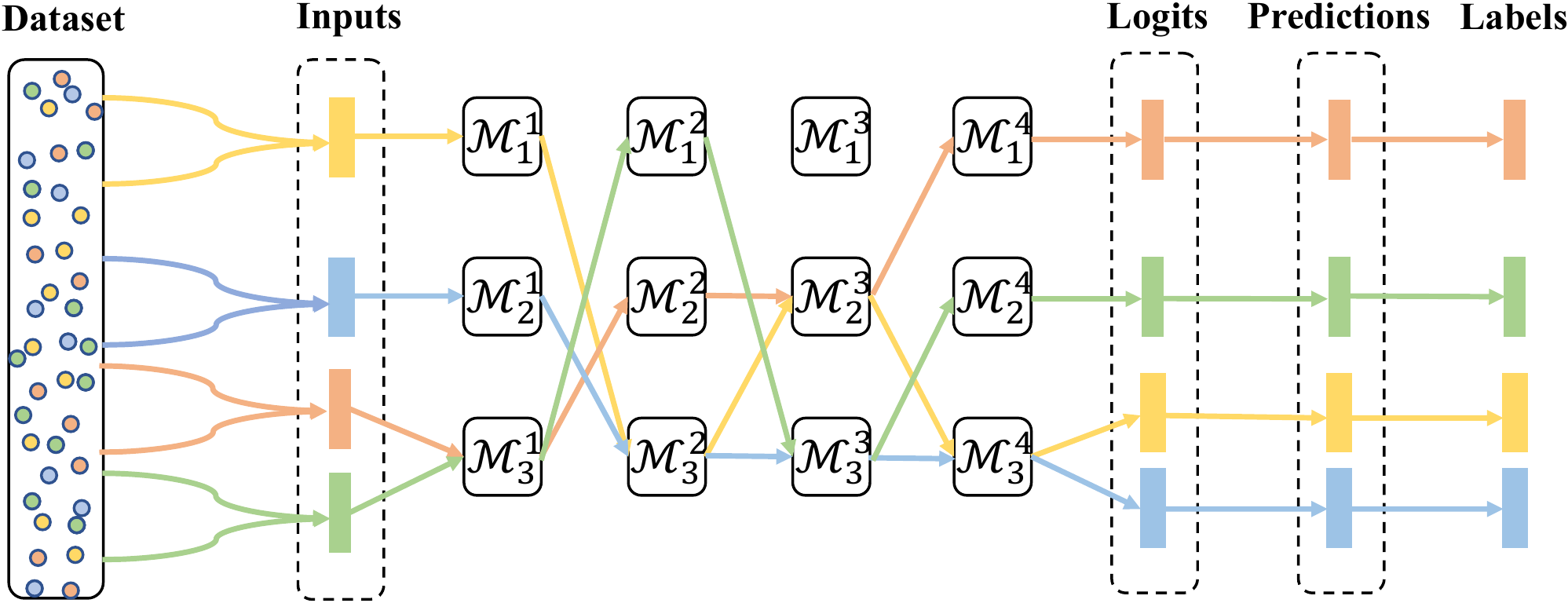}
    \caption{\textit{An overview of collaborative group learning with random routing and sub-set data learning.}}
    \label{fig:Random path selection}
    % \vspace{-1.0\baselineskip}
\end{figure}

Previous work constructs a pool of students by continually introducing new networks or branches, which rapidly expands the model capacity with the increasing number of students. In our proposed collaborative learning, students are built from a modular neural network, using random routing, where students can be built as many as possible given restricted-capacity networks. More importantly, random module sharing and branching also benefit knowledge interaction between students. The modular neural network consists of $L$ distinct layers with each layer $\ell \in[1, L]$ containing $M$ modules, arranged in parallel, i.e., $\mathcal{M}^{\ell}=\left\{\mathcal{M}_{m}^{\ell}\right\}_{m=1}^{M}$ (see Figure \ref{fig:Random path selection}). Each module $\mathcal{M}_{m}^{\ell}$ is a learnable sub-network embedded in the modular neural network, consisting of different combinations of layers. It extracts different types of features in accordance with various tasks, such as residual block for vision features and transformer block for semantic features. For the $\ell^{th}$ layer, the index of the selected module is uniformly sampled using $\mathbb{U}(\cdot)$ over the set of integers $[1,M]$. After $L$ times of selection, we construct a pathway $\mathbf{P}_{k} \in \mathbb{R}^{L \times M}$ to form a random routing network for the $k^{th}$ student:
\begin{equation}
\small
\mathbf{P}_{k}(\ell, m) = \left\{\begin{array}{ll}1, & \text { if the module } \mathcal{M}_{m}^{\ell} \text { is present in the path, } \\ 0, & \text { otherwise. }\end{array}\right.
\end{equation}
When training the $k^{th}$ student, any $m^{th}$ module in $\ell^{th}$ layer with $\mathbf{P}_{k}(\ell, m)=1$ is activated. All established students are simultaneously trained by two supervised losses that we will elaborate later. With the help of random routing, one student can share the selected module of any other student in the same layer, which means that compared with previous alternative work, our method can build more students with the same number of parameters. Also, more students participating in collaboration imply more diverse feature sets in the student pool. Meanwhile, the fine-grained representation sharing and branching across multi-layers between students, as an extreme form of hint training \cite{DBLP:journals/corr/RomeroBKCGB14}, implicitly and flexibly boost knowledge sharing and transfer.
Consequently, it naturally imposes efficient regularization on the feature learning for each student \cite{DBLP:conf/nips/SongC18,DBLP:conf/nips/LanZG18}.

%%%%%%%%%%%%%%%%%%%%%%%%%%%%%%%%%%%%%%%%%%%%%%%%%%%%%%%%%%%%%%%%%%%%%%%%%%%%%%%%%%
%%%%%%%%%%%%%%%%%%%%%%%%%%%%%%%%%%%%%%%%%%%%%%%%%%%%%%%%%%%%%%%%%%%%%%%%%%%%%%%%%%

\subsection{Sub-set Data Learning}
In prior collaborative approaches, all students learn from the same set of training data. The inductive bias contained in the training data significantly affects the features learning \cite{DBLP:conf/aaai/ZhangWZ18}, facilitates or hinders the model training \cite{DBLP:conf/icml/SuchRLSC20}. We consider utilizing the inductive bias of training data to enrich the diversity of students and propose a sub-set data learning. Concretely, given $N$ samples $\mathcal{X}=\left\{x_{i}\right\}_{i=1}^{N}$ from $C$ classes with the corresponding labels $\mathcal{Y}=\left\{y_{i}\right\}_{i=1}^{N}$, where $y_{i}\in\{1, 2, \ldots, C\}$, the entire training set is randomly divided into $K$ subsets $\mathcal{X}^{k}=\left\{x_{i}^{k}\right\}_{i=1}^{N/K}$ to train the corresponding $K$ students (see Figure \ref{fig:Random path selection}). The $k^{th}$ student produces the probability of class $c$ for sample $x_{i}^{k}$ by normalizing the logit,
\begin{equation}
\small
p_{k}^{c}\left(x_{i}^{k}\right) = \frac{\exp \left(z_{k}^{c}\right)}{\sum_{j=1}^{C} \exp \left(z_{k}^{j}\right)}
\end{equation}
where the logit $z_{k}$ is the output of the $k^{th}$ student. 

As a multi-class classifier, the general training criterion of the $k^{th}$ student is to minimize the cross-entropy between the ground-truth labels and the predicted distributions,
\begin{equation}
\small
L_{ce}^{k}=-\sum_{i=1}^{N/K} \sum_{c=1}^{C} \mathbb I\left(y_{i}^{k} = c\right) \log \left(p_{k}^{c}\left(x_{i}^{k}\right)\right)
\end{equation}
where $\mathbb I\{\cdot\}$ is the indicator function.

\subsection{Sub-group Imitation}
\begin{figure}[t]
    \centering
    \includegraphics[width=0.40\linewidth]{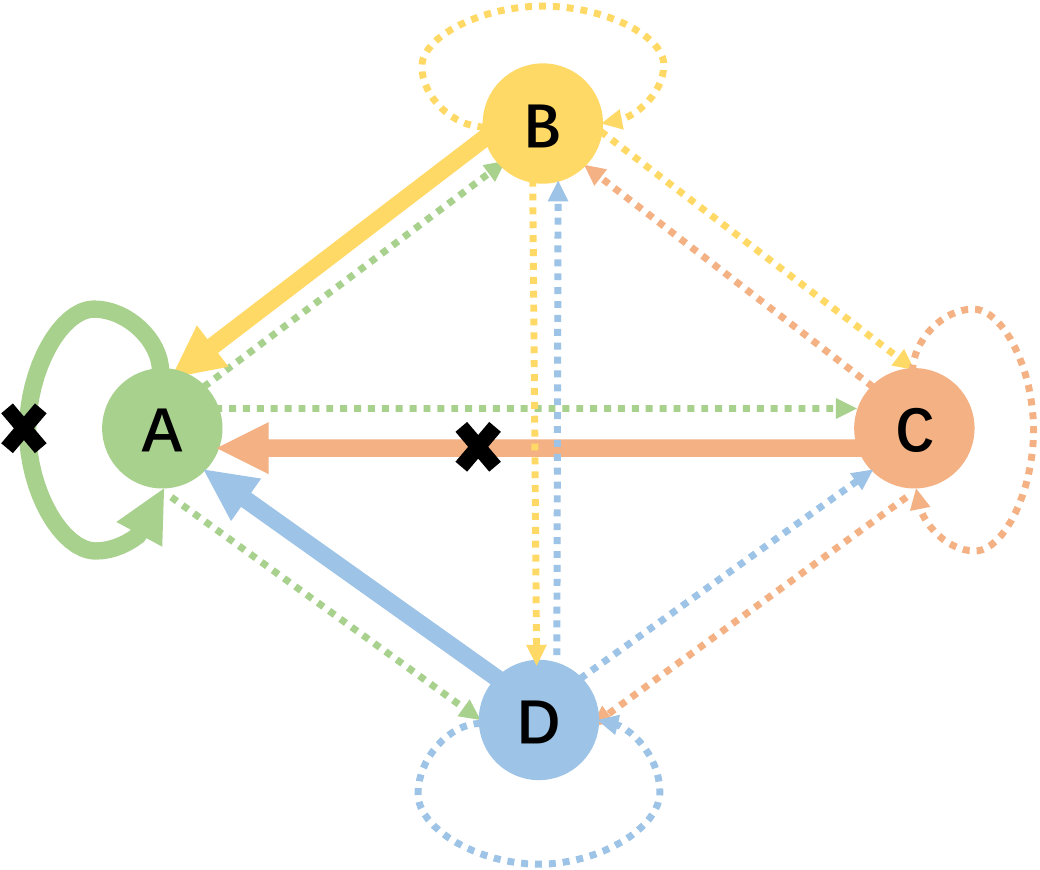}
    \caption{\textit{Sub-group imitation}. For example, with the imitating probability $p=0.5$, student A may only choose student B and D to aggregate the group-level knowledge for one iteration.}
    \label{fig:random collaborative strategy}
    % \vspace{-1.0\baselineskip}
\end{figure}

Conventional collaborative learning usually introduces extra supplementary information in the form of group-level knowledge. However, aggregating group-level knowledge with the naive or the weighted average faces two main drawbacks: First, the homogenization phenomenon is more likely to occur due to the similar and redundant group-level knowledge of students; Besides, the computational cost increase linearly as the number of students continually grows. In order to improve the generalization of each student, we propose a sub-group imitation (see Figure \ref{fig:random collaborative strategy}), which randomly selects a sub-group instead of the whole group of students for imitating in each iteration. Intuitively, in our collaborative framework, each student follows a dynamic and evolving ``teacher'' to gain experiences while learns to denoise the random perturbation of soft knowledge (prediction alignment) and hard knowledge (parameter sharing) that alleviates student homogenization but hinders the stability of student learning. In practice, we can adjust the sub-group size flexibly to balance the performance and training computational cost. The group-level knowledge for the $k^{th}$ student is computed as:
\begin{equation}
\small 
\begin{aligned}
& L_{kl}^{k} =  
             \sum_{i=1}^{N/K} \sum_{c=1}^{C}
             p_{t}^{c} \left(x_{i}^{k}; T\right) 
             \log \frac{
             p_{t}^{c}\left(x_{i}^{k}; T\right)}
             {p_{k}^{c}\left(x_{i}^{k}; T\right)} 
             ; \quad \\
& p_{t}^{c} \left( x_{i}^{k}; T \right) = 
            \frac{\exp \left( z_{t}^{c} /T \right)}
                 {\sum_{j=1}^{C} \exp \left( z_{t}^{j}/T \right)}
\end{aligned}
\end{equation}
where $T$ is the temperature, used to soften the predictions, and $z_{t}$ is defined as:
\begin{equation}
\small
z_{t} = \frac{1}{H} \sum_{k=1}^{K} Select \left( z_{k} \right)
\end{equation}
where $H$ is the expected number of imitated students, and $Select(\cdot)$ is the selection function with an imitating probability $p$. Note that we first select which students to imitate and then calculate the corresponding $z_{k}$ in practice.

%%%%%%%%%%%%%%%%%%%%%%%%%%%%%%%%%%%%%%%%%%%%%%%%%%%%%%%%%%%%%%%%%%%%%%%%%%%%%%%%%%
%%%%%%%%%%%%%%%%%%%%%%%%%%%%%%%%%%%%%%%%%%%%%%%%%%%%%%%%%%%%%%%%%%%%%%%%%%%%%%%%%%

\subsection{Optimization}
We obtain the overall loss function as:
\begin{equation}
\small
L = \sum_{k=1}^{K} \left( L_{ce}^{k} + \phi(t) * L_{kl}^{k} \right),
\end{equation}
where $\phi(t)$ is a ramp-up coefficient function \cite{DBLP:conf/iclr/LaineA17} that maintains an equilibrium of the contribution of labels and group-level knowledge. The imbalance of contribution will result in either exacerbating the homogenization of students or weaken knowledge transfer between students. The ramp-up coefficient function can prevent students from getting prematurely stuck in the homogenization problem, which causes that students can not learn enough diverse knowledge to regularize each other effectively.
\begin{equation}
\small
\phi(t)= 
\left \{ 
\begin{array}{ll} 
1, & \text{ if $t$ not in } [J^{s}, J^{e}], \\ 
\exp \left( -5*(1-\lambda)^{2} \right), & \text{ otherwise. }
\end{array}
\right .
\end{equation}
where $t$ is the index of training epoch, and $\lambda$ is a scalar that increases linearly from zero to one during the ramp-up range $[J^{s}, J^{e}]$. Once a pool of students are collected from the proposed modular neural network, and the training set is randomly divided, we conduct the sub-group imitation throughout the whole training process. All students are trained simultaneously at each iteration until convergence. In inference, we can randomly select one student or choose the best student by a hold-out set to predict the class of input data.
\section{Experiments}

\begin{table*}[!t]
\scriptsize
\centering
\begin{tabular}{@{}lrrrrl@{}}
\hline
Dataset & \multicolumn{1}{l}{\# Train} & \multicolumn{1}{l}{\# Holdout} & \multicolumn{1}{l}{\# Test} & \multicolumn{1}{l}{\# Classes} & Classification Task \\ \hline
CIFAR-10 \cite{krizhevsky2009learning} & 45k & 5k & 10k & 10 & Image classification \\
CIFAR-100 \cite{krizhevsky2009learning} & 45k & 5k & 10k & 100 & Image classification \\
IMDB Review \cite{DBLP:conf/acl/MaasDPHNP11} & 23k & 2k & 25k & 2 & Sentiment analysis \\
Yelp Review Full \cite{DBLP:conf/nips/ZhangZL15} & 630k & 20k & 50k & 5 & Sentiment analysis \\
Yahoo! Answers \cite{DBLP:conf/nips/ZhangZL15} & 1 350k & 50k & 60k & 10 & Topic classification \\
Amazon Review Full \cite{DBLP:conf/nips/ZhangZL15} & 2 900k & 100k & 650k & 5 & Sentiment analysis \\ \hline
\end{tabular}
\caption{Statistics of six classification datasets used in our experiments.}
% \vspace{-1.0\baselineskip}
\label{tab: datasets description}
\end{table*}

\begin{table*}[t]
\scriptsize
\centering
\setlength{\tabcolsep}{4.0pt}
\begin{tabular}{lcccccc}
\hline
Datasets & Baseline & DML & CL-ILR & ONE & \multicolumn{1}{c|}{OKDDip} & CGL \\ \hline
\multicolumn{7}{c}{ResNet-18} \\ \hline
CIFAR-10 & $93.97 \pm 0.09$ & $94.18 \pm 0.09$ & $94.11 \pm 0.12$ & $94.19 \pm 0.06$ & \multicolumn{1}{c|}{$94.29 \pm 0.04$} & \boldsymbol{$94.61 \pm 0.06$} \\
CIFAR-100 & $74.68 \pm 0.13$ & $76.13 \pm 0.10$ & $76.61 \pm 0.03$ & $76.17 \pm 0.12$ & \multicolumn{1}{c|}{$76.69 \pm 0.04$} & \boldsymbol{$78.01 \pm 0.07$} \\ \hline
\multicolumn{7}{c}{ResNet-34} \\ \hline
CIFAR-100 & $76.06 \pm 0.11$ & $76.73 \pm 0.12$ & $77.09 \pm 0.12$ & $76.96 \pm 0.10$ & \multicolumn{1}{c|}{$77.39 \pm 0.09$} & \boldsymbol{$78.31 \pm 0.10$} \\ \hline
\end{tabular}
\caption{Top-1 accuracy (\%) on the image datasets.}
% \vspace{-1.0\baselineskip}
\label{tab: results on image classification}
\end{table*}

\subsection{Datasets and Architectures} 
We present our results on six public available datasets of three classification tasks covering image classification, topic classification, and sentiment analysis. To validate the effectiveness of the proposed collaborative group learning framework in depth, we conduct the evaluation tasks in various tasks ranging from image field to more challenging text classifications, especially the fine-grained sentiment analysis tasks. Table \ref{tab: datasets description} summarizes the statistics of all datasets. For the image-related tasks, we adopt augmentation and normalization procedure following \cite{DBLP:conf/cvpr/HeZRS16}. 
For the text-related tasks, following \cite{DBLP:conf/eacl/SchwenkBCL17}, we do not conduct any preprocessing except lower-casing. Four network architectures are used in our experiments for different tasks, ResNet-18 and ResNet-34 \cite{DBLP:conf/cvpr/HeZRS16} for CIFAR-10 and CIFAR-100, Transformer \cite{DBLP:conf/nips/VaswaniSPUJGKP17} for IMDB Review, and VDCNN-9 \cite{DBLP:conf/eacl/SchwenkBCL17} for the rest of datasets.

\subsection{Comparison Approaches}
We compare \textbf{C}ollaborative \textbf{G}roup \textbf{L}earning (CGL) to several recently proposed collaborative approaches, including network-based DML \cite{DBLP:conf/cvpr/ZhangXHL18}, branch-based CL-ILR \cite{DBLP:conf/nips/SongC18}, ONE \cite{DBLP:conf/nips/LanZG18}, and OKDDip \cite{DBLP:conf/aaai/ChenMWF020}. We also report a “Baseline” model that trains only one student on ground-truth labels. For branch-based approaches, all students share the first several blocks of layers and separate from the last block to form a multi-branch structure as \cite{DBLP:conf/nips/LanZG18}. The students in all the comparison models are set to the same amount and architecture for different tasks, i.e., 3 students for image datasets and 5 students for text datasets. The number of parameters increases as more students join in, whereas in our method, given 9 layers of modules and 2 modules in each layer, with random routing, theoretically we can build 2 to 512 different students without extra computational cost. In our experiment, we set 8 students for collaborative group learning. The imitating probability is set to 0.25 for image tasks and 0.5 for text tasks. The student that obtains the best score on the holdout set is used for evaluation. In OKDDip \cite{DBLP:conf/aaai/ChenMWF020}, the group leader student is chosen for prediction.

\subsection{Experiment Settings}
For ResNet-18 and ResNet-34, we use Adam \cite{DBLP:journals/corr/KingmaB14} for optimization with a mini-batch of size 64. The initial learning rate is 0.001, divided by 2 at 60, 120, and 160 of the total 200 training epochs. For VDCNN-9, we adopted the same experimental settings as \cite{DBLP:conf/eacl/SchwenkBCL17,DBLP:conf/nips/ZhangZL15}. Training is performed with Adam, using a mini-batch of size 64, a learning rate of 0.001 for the total 20 training epochs. We use SentencePiece\footnote{https://github.com/google/sentencepiece} (BPE) to tokenize IMDB Review and set vocabulary size, embedding dimension, and maximum sequence length to 16000, 512, and 512. For Transformer, the size of blocks and heads is 3 and 4 separately. We set the size of the hidden state and feed-forward layer to 128 and 512. Training is performed with Adam, using a mini-batch of size 64, a learning rate of 0.0001 for the total 30 training epochs. We run each method 3 times and report ``mean (std)''.

\subsection{Comparison on Image Classification}
Table \ref{tab: results on image classification} summarises the Top-1 accuracy (\%) of CIFAR-10 and CIFAR-100 obtained by ResNet-18 and ResNet-34 with the existing state-of-the-art and our methods. We observe that our method significantly outperforms all other methods with substantial accuracy gains, which shows that with the same computational cost, our collaborative framework is more effective than previous methods on improving model generalization. The branch-based methods, especially OKDDip, yield more generalizable models compared to the network-based method (DML). This suggests parameter sharing benefits the transfer of diverse and complementary knowledge between students as observations in \cite{DBLP:conf/nips/SongC18,DBLP:conf/nips/LanZG18}. We also found that all collaborative frameworks achieve more performance improvement in the smaller architecture according to the results of ResNet-18 and ResNet-34 on CIFAR-100.

\begin{table*}[t]
\scriptsize
\centering
\setlength{\tabcolsep}{4.0pt}
\begin{tabular}{lcccccc}
\hline
Datasets & Baseline & DML & CL-ILR & ONE & \multicolumn{1}{c|}{OKDDip} & CGL \\ \hline
\multicolumn{7}{c}{VDCNN-9} \\ \hline
Yelp Review Full & $62.15 \pm 0.15$ & $62.53 \pm 0.10$ & $62.66 \pm 0.08$ & $62.74 \pm 0.05$ & \multicolumn{1}{c|}{$62.75 \pm 0.18$} & \boldsymbol{$63.32 \pm 0.04$} \\
Yahoo! Answers & $69.02 \pm 0.07$ & $69.79 \pm 0.11$ & $70.09 \pm 0.07$ & $70.08 \pm 0.09$ & \multicolumn{1}{c|}{$70.10 \pm 0.09$} & \boldsymbol{$70.35 \pm 0.05$} \\
Amazon Review Full & $60.25 \pm 0.11$ & $60.54 \pm 0.10$ & $60.59 \pm 0.07$ & $60.49 \pm 0.03$ & \multicolumn{1}{c|}{$60.63 \pm 0.04$} & \boldsymbol{$61.03 \pm 0.05$} \\ \hline
\multicolumn{7}{c}{Transformer} \\ \hline
IMDB Review & $82.30 \pm 0.10$ & $82.45 \pm 0.05$ & $83.10 \pm 0.07$ & $82.66 \pm 0.08$ & \multicolumn{1}{c|}{$82.74 \pm 0.12$} & \boldsymbol{$83.81 \pm 0.10$} \\ \hline
\end{tabular}
\caption{Top-1 accuracy (\%) on the text datasets.}
% \vspace{-1.0\baselineskip}
\label{tab: results on text classification}
\end{table*}

\subsection{Comparison on Text Classification}
Table \ref{tab: results on text classification} reports the Top-1 accuracy (\%) of all text datasets based on VDCNN-9 and Transformer. It can be seen that our method also achieves better performance than all prior methods as above, indicating that our method can be generically applied to more challenging text classification tasks. The prior methods obtain slightly better performance than ``Baseline'' on all datasets except for Yahoo! Answers (Topic classification), which means that the difficulty of clearly discriminate fine-grained sentiment labels hinders students from discovering diverse feature sets and transferring supplementary knowledge from the others. The superiority of our method on both image and text datasets demonstrates the generalization and robustness of the proposed collaborative framework.

\subsection{Ablation Study and Analysis}
In this section, we further investigate the effectiveness and robustness of our method, including random routing, sub-set data learning, and sub-group imitation. We also provide detailed analyses to demonstrate how and why our method works. We conduct ablation comparisons with the branch-based approaches, as they have the advantages of better performance and lower computational cost. The score reported below is all obtained by running each model 3 times and providing ``mean''. Unless otherwise stated, the following results are based on CIFAR-100 with ResNet-18.

\paragraph{Ablation Study} 
In Table \ref{tab: ablation study}, we report the Top-1 accuracy (\%) of models ``w/o random routing (RR)'' (i.e., built by individual networks), ``w/o sub-set data learning (SDL)'' (i.e., using the same entire training data), and ``w/o sub-group imitation (SGI)'' (i.e., imitating all other students). According to the results, we can see that 1) without the parameter sharing generated by random routing, students, only based on the assigned sub-set data and the logits-based imitation, do not obtain sufficient information for the feature learning, which demonstrates that random routing indeed brings a highly effective knowledge transfer. 2) without sub-set data learning or sub-group imitation, the student homogenization will aggravate and the model may converge to a worse sub-optimal. These phenomena verify that the reason that CGL works well is the \textbf{\textit{effective balance}} of conducting knowledge transfer and maintaining the model diversity. 

To analyze the impact of the ramp-up coefficient, we set the ramp-up interval to 0\%, 20\%, 40\% and 80\% of the training epochs. The results of Table \ref{tab: ramp-up} indicate the ramp-up coefficient can alleviate the homogenization problem but too long ramp-up ranges weaken knowledge transfer.

\begin{table}
\scriptsize
\centering
\setlength{\tabcolsep}{7.5pt}
\begin{tabular}{lccc}
\hline
Condition & w/o RR & w/o SDL & w/o SGI \\ \hline
Accuracy (\%) & 75.75 & 77.48 & 77.23 \\ \hline
\end{tabular}
\caption{Results of the ablation study.}
% \vspace{-1.0\baselineskip}
\label{tab: ablation study}
\end{table}

\begin{table}
\scriptsize
\centering
\setlength{\tabcolsep}{7.0pt}
\begin{tabular}{lcccc}
\hline
Ramp-up Range (\%) & 0 & 20 & 40 & 80 \\ \hline
Accuracy (\%) & 77.79 & 78.01 & 77.22 & 77.10\\ \hline
\end{tabular}
\caption{Impact of the Ramp-up coefficient.}
% \vspace{-1.0\baselineskip}
\label{tab: ramp-up}
\end{table}

\begin{figure}[t]
\centering
\vfil
\subcaptionbox{Comparative performances\label{fig: number of agents, accuracy}}{ \includegraphics[width=0.50\linewidth]{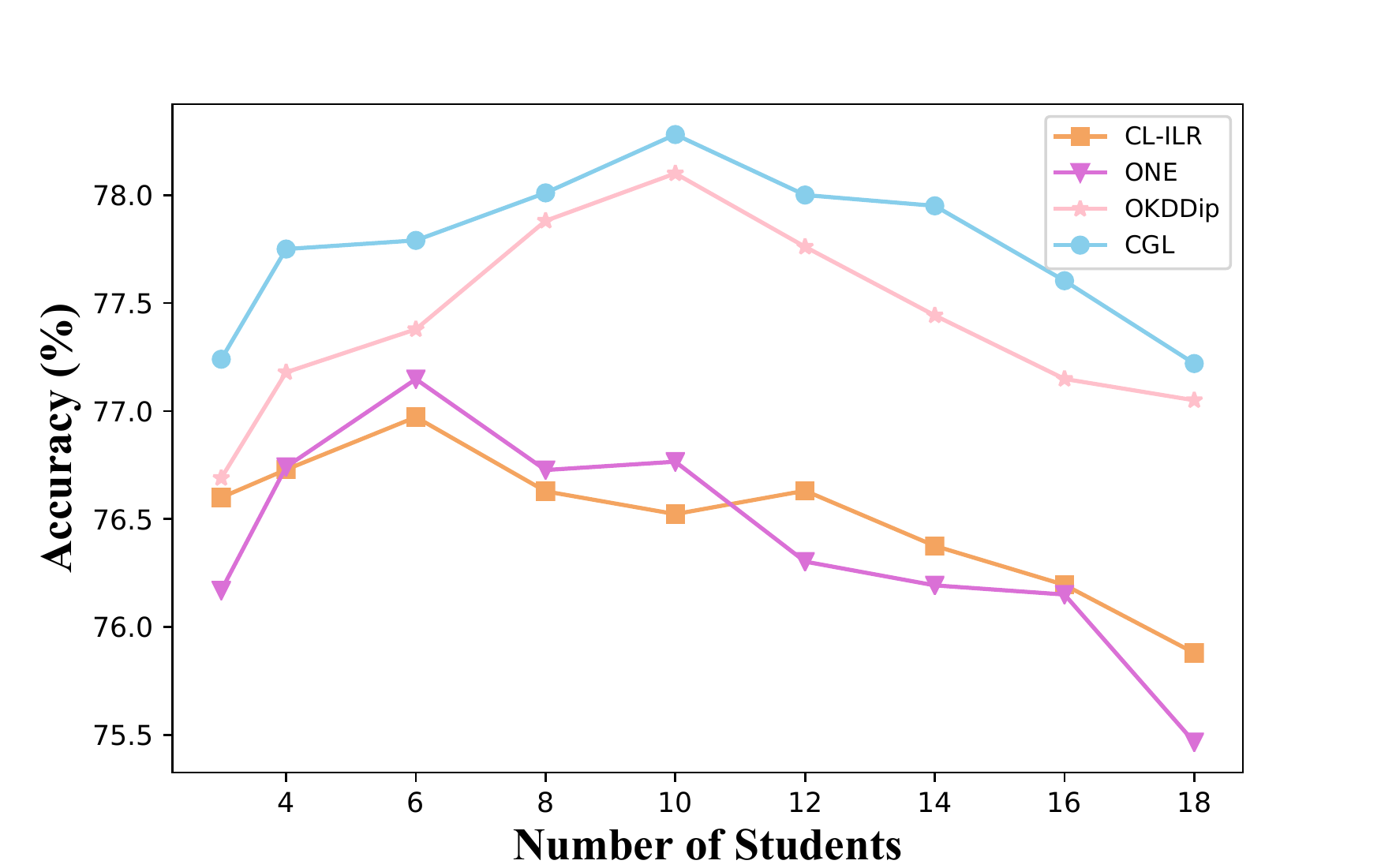}}
\vfil
\subcaptionbox{Number of parameters\label{fig: number of agents, parameter}}{ \includegraphics[width=0.50\linewidth]{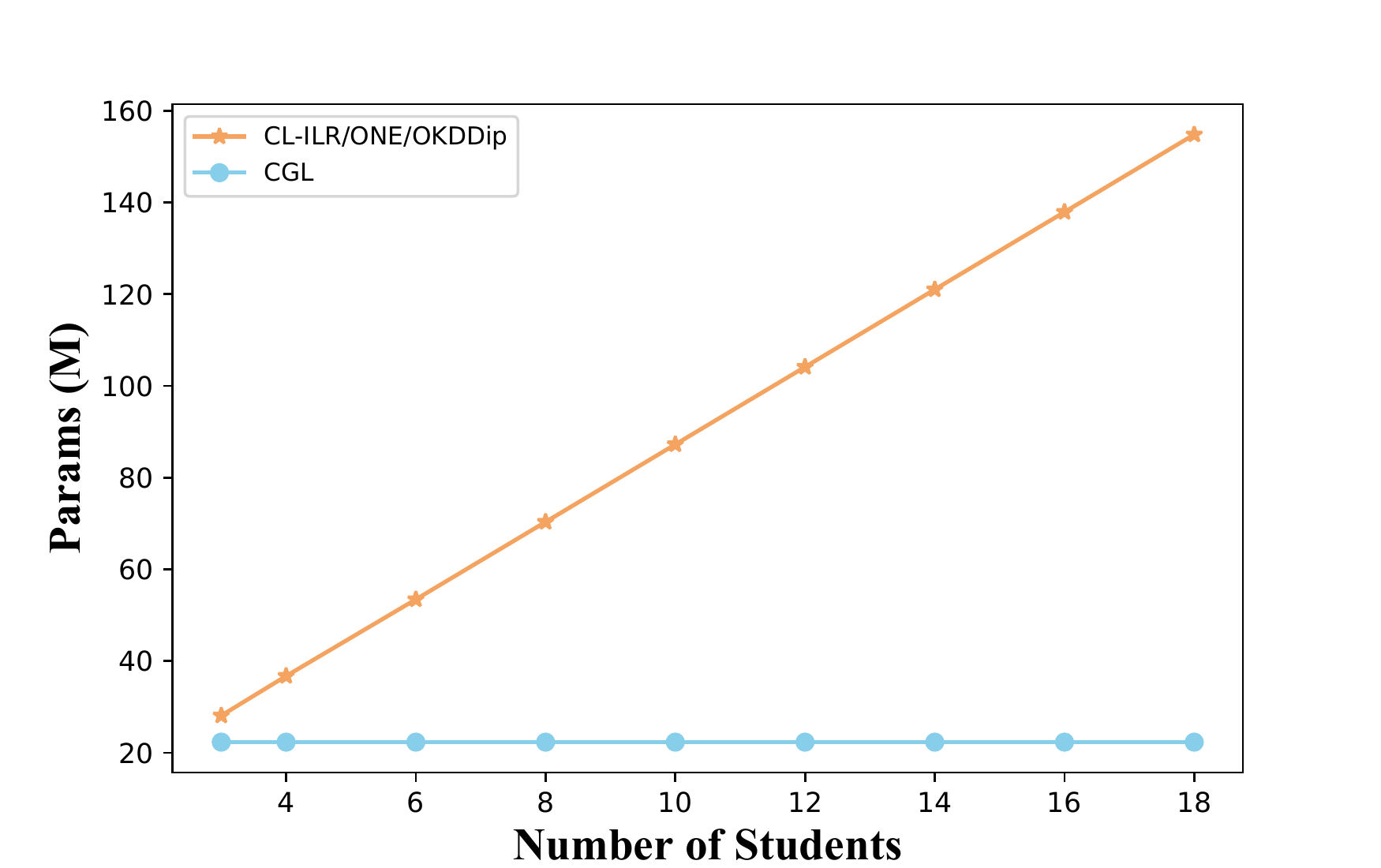} }
\vfil
	\caption{Impact of student population. The parameter number of comparative methods explodes as more students are involved, while our method maintains a constant computational cost.} 
	\label{fig: number of agents} 
	% \vspace{-1.0\baselineskip}
\end{figure}

\paragraph{Impact of Student Population}
It is well known that the increasing number of students benefits the model performance \cite{DBLP:conf/nips/LanZG18,DBLP:conf/aaai/ChenMWF020}. Figure~\ref{fig: number of agents}(\subref{fig: number of agents, accuracy}) shows the Top-1 accuracy (\%) of all comparative methods with respect to the number of students. Our method consistently achieves the best accuracy in varying numbers of students, which demonstrates its superiority. We also observe that the curves of all methods rise first and then decline, which implies that excessive students also damage the model performance. We conjecture that excessive students may not be sufficiently trained due to weakening knowledge discovery of each student (for our methods) or exacerbating the similarity and redundancy of group-level knowledge (for comparative methods). In such a scenario, our method is undergoing under-fitting, whereas comparative methods are usually struggling with over-fitting (too many students homogenizing). Such an under-fitting problem can be simply adjusted by allowing sub-set data overlapping or increasing imitation probability. The performance of CI-ILR, ONE declines much earlier than OKDDip and CGL, which indicates that the latter effectively handles more diverse students.
 
However, more students significantly increase the number of model parameters and the computational cost of training (i.e., \textbf{computational efficiency}), which limits the deployment of collaborative learning. Our method can alleviate these problems by random routing and sub-group imitation. In terms of model parameters, as shown in Figure~\ref{fig: number of agents}(\subref{fig: number of agents, parameter}), our method remains a constant number of model parameters as more students join in, whereas the parameter number in comparison models explodes. Moreover, our method also maintains constant computational cost (the number of forward and backward propagation) by variable imitation probability. As for the training time, similar to codistillation (a variant of DML) \cite{DBLP:conf/iclr/AnilPPODH18}, our method can be easily implemented in parallel. Therefore it still consumes the constant training time as the number of students increases. Please refer to the appendix for a more detailed discussion.

%%%%%%%%%%%%%%%%%%%%%%%%%%%%%%%%%%%%%%%%%%%%%%%%%%%%%%%%%%%%%%%%%%%%%%%%%%%%%%%%%%
\begin{figure}[t]
\centering
\includegraphics[width=0.65\linewidth]{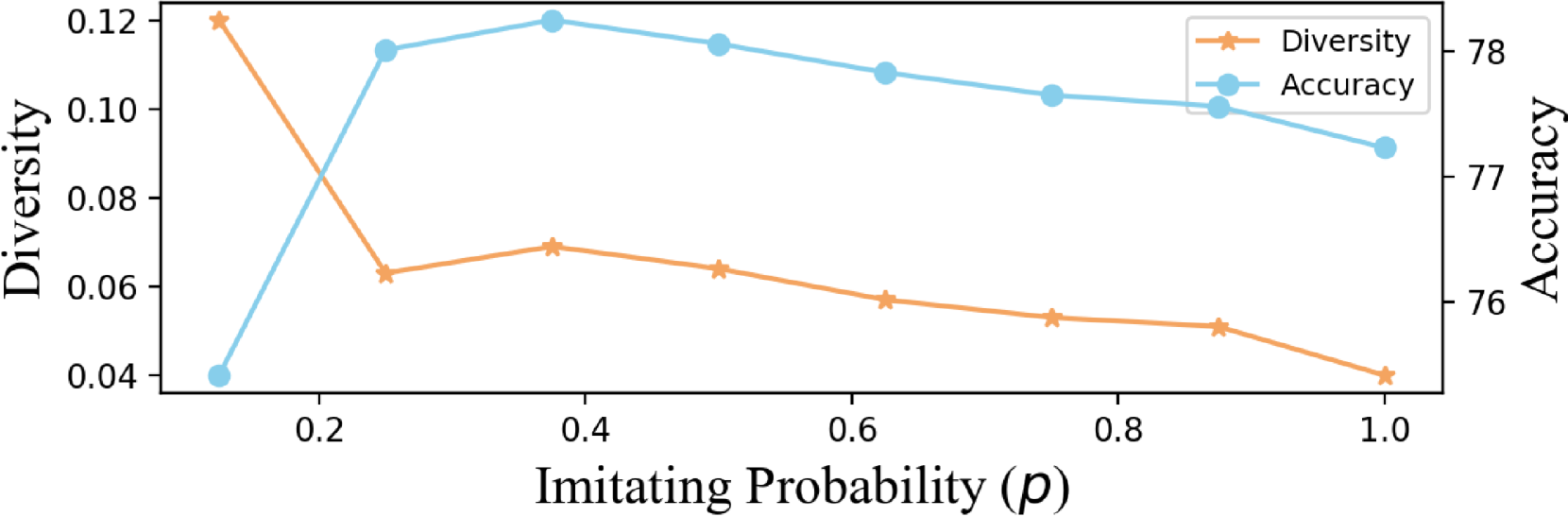}
\caption{Effect of imitating probability.}
% \vspace{-1.0\baselineskip}
\label{tab: Diversity; imitation path}
\end{figure}
%%%%%%%%%%%%%%%%%%%%%%%%%%%%%%%%%%%%%%%%%%%%%%%%%%%%%%%%%%%%%%%%%%%%%%%%%%%%%%%%%%

\begin{table}[t]
\scriptsize
\centering
\setlength{\tabcolsep}{4.5pt}
\begin{tabular}{lll}
\hline
Dataset (Architecture) & \multicolumn{1}{c}{Condition} & \multicolumn{1}{c}{Diversity} \\ \hline
\multirow{2}{*}{CIFAR-100 (ResNet-18)} & (w/.) & 0.535 \\
& (w/o.) & 0.178 \\ \hline
\multirow{2}{*}{Yelp Review Full (VDCNN-9)} & (w/.) & 0.185 \\
& (w/o.) & 0.131 \\ \hline
\multirow{2}{*}{IMDB Review (Transformer)} & (w/.) & 0.056 \\
& (w/o.) & 0.039 \\ \hline
\end{tabular}
\caption{Effect of sub-set data learning.}
% \vspace{-2.0\baselineskip}
\label{tab: Diversity w/. subset and w/o. subset}
\end{table}

\paragraph{Model Diversity Analysis}
We improve the model diversity in collaborative learning from two aspects: sub-set data learning diversifies the training sets of students to learn diverse feature sets; sub-group imitation randomly selects various sub-groups of imitated students from which students aggregate and distill different complementary knowledge. Table \ref{tab: Diversity w/. subset and w/o. subset} reports the diversity of students $w.$ and $w/o.$ sub-set data learning based on a comprehensive set of architectures and datasets. The diversity is calculated by averaging $L_{2}$ distance between the probability distribution of each pair of students. To isolate the effect of sub-set data learning, we revoke sub-group imitation in this analysis. The results in Table \ref{tab: Diversity w/. subset and w/o. subset} verify sub-set data learning indeed boosts the diversity of students on various architectures and datasets. 

We further vary the imitating probability $p$ to analyze its effect on diversity and accuracy. The imitating probability $p$ is selected from $[0.125, 1.0]$ and applied to ResNet-18 on CIFAR-100. From Figure \ref{tab: Diversity; imitation path}, we discover that the diversity shows a downward trend, and the accuracy first ascends and then slowly declines. This phenomenon demonstrates that when each student aggregates knowledge from too many students, the model performance declines as they may homogenize to each other; however, when it mimics very few students, it is unable to distill a sufficient amount of knowledge. Our method achieves a better balance between diversity and performance with limited computational resources by choosing a proper imitating probability.

\begin{figure}[t]
\centering
\vfil
\subcaptionbox{Autonomous parameter sharing\label{fig: parameter sharing, reducing}}{\includegraphics[width=0.65\linewidth]{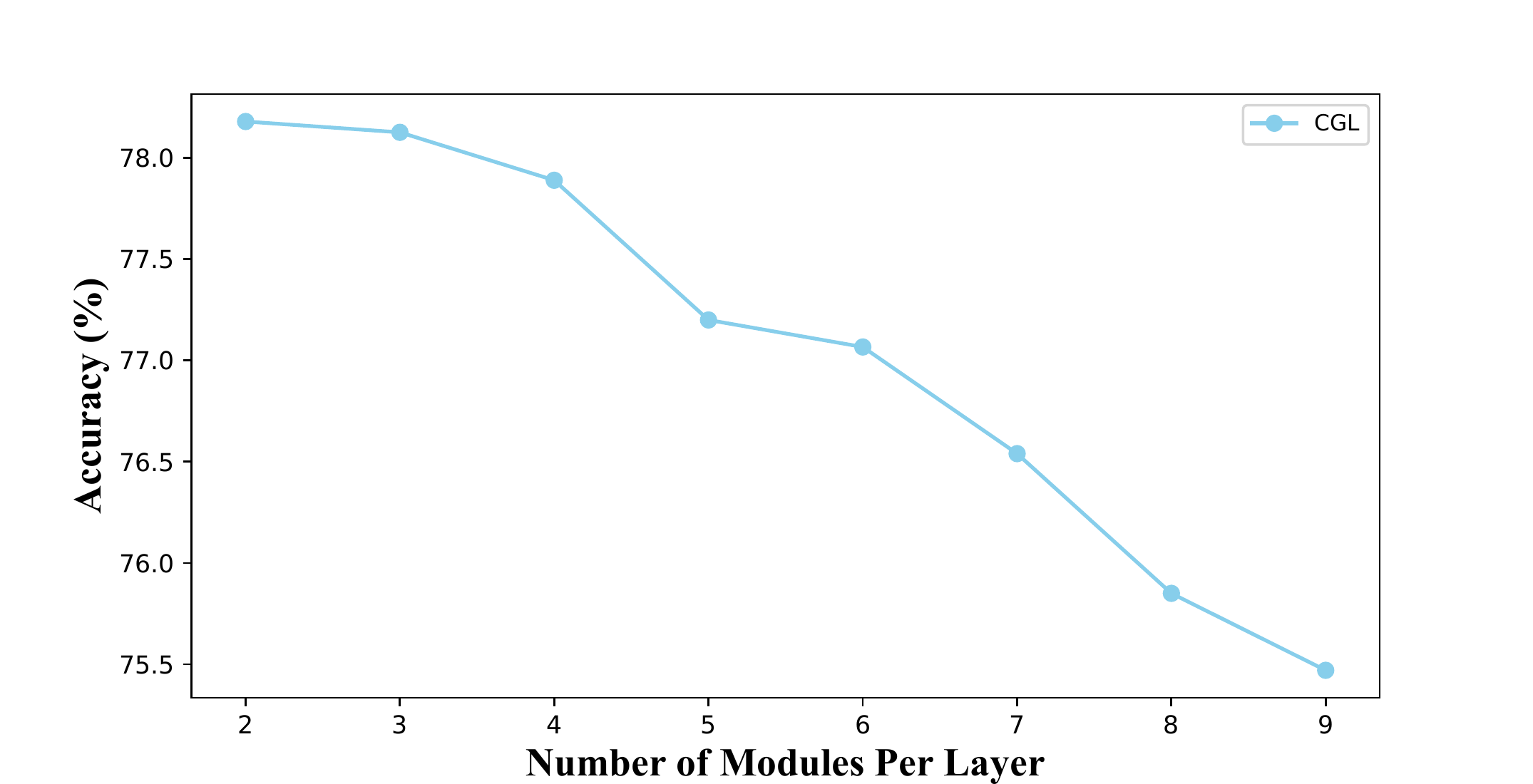}}
\vfil
\subcaptionbox{Enforced parameter sharing\label{fig: parameter sharing, increasing}  }{\includegraphics[width=0.65\linewidth]{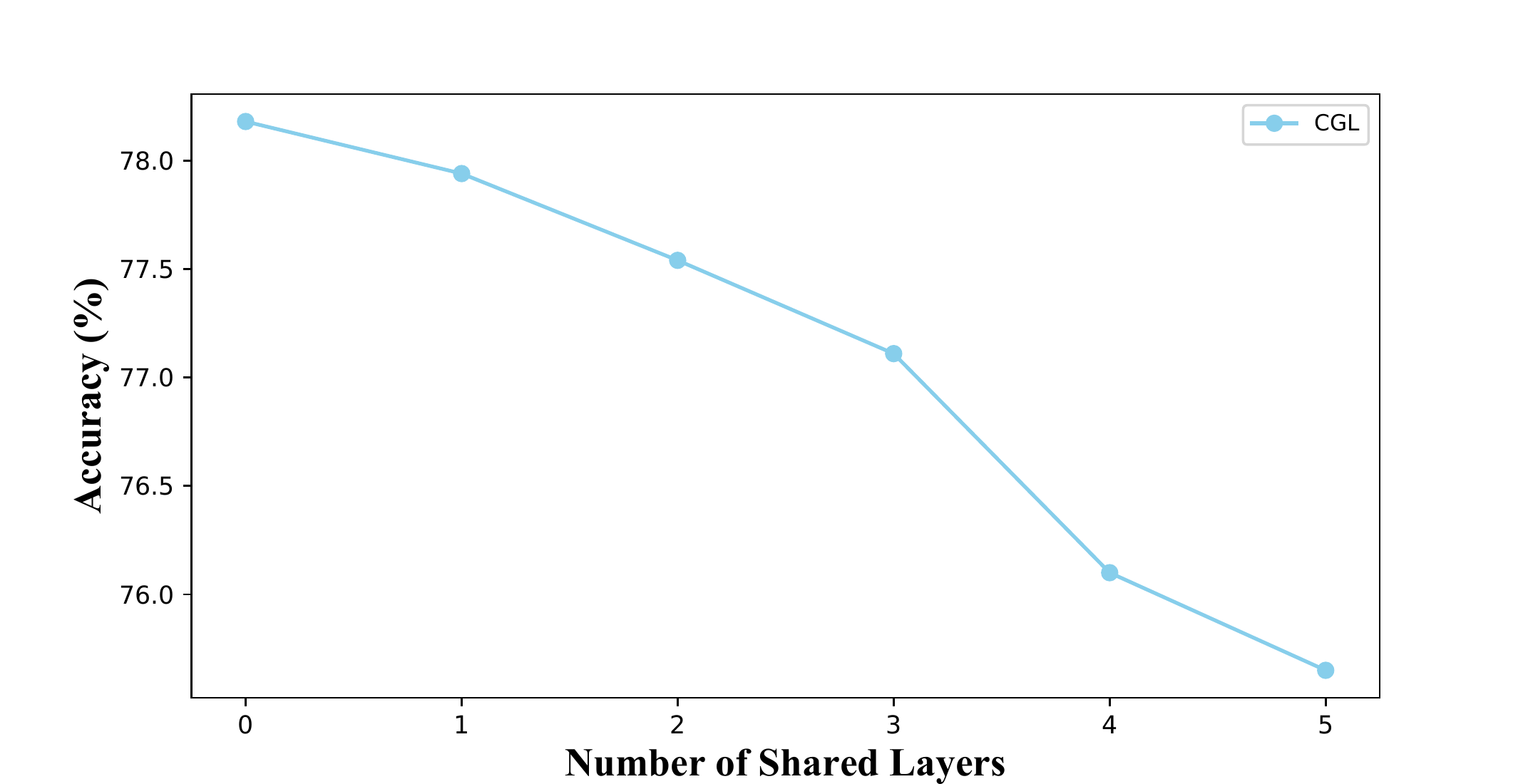}}
\vfil
	\caption{Impact of parameter sharing ratio. We can sparse parameter sharing by increasing the number of modules per layer or densify parameter sharing by manually setting more shared layers in which all students choose the same module.} 
    % \vspace{-1.0\baselineskip}
	\label{fig: parameter sharing}  
\end{figure}

\paragraph{Impact of Parameter Sharing}
Besides considering the logits of the output layer, parameter sharing is also an implicit and efficient way to boost knowledge transfer by aligning the intermediate features between selected students \cite{DBLP:conf/nips/SongC18,DBLP:conf/nips/LanZG18}. Network-based collaborative learning does not support parameter sharing, while branch-based one shares parameters only at the bottom layers. Benefited from random routing, our method naturally allows flexible knowledge communication based on fine-grained and random levels of parameter sharing structure. We first investigate the effect of parameter sharing ratio on model performance. We fix the number of students and the imitating probability, and then vary the parameter sharing ratio by adjusting the number of modules per layer or manually setting shared layers. From Figure \ref{fig: parameter sharing}, we discover that for collaborative learning, sharing too many layers will cause students to homogenize, and sharing too few layers weaken knowledge transfer, which implies previous parameter sharing structure is not flexible enough to maintain a trade-off between diversity and generalization of students due to dense and consecutive multi-layer parameter sharing.

Similar to neural architecture search \cite{DBLP:conf/iclr/ZophL17}, our collaborative framework is capable of finding an efficient and generalizable parameter sharing structure. Concretely, one can collect a set of parameter sharing structures by random routing, and then select the best structure that can be applied directly to a new dataset. We validate this assumption by randomly generating eight parameter sharing structures on CIFAR-10 based on ResNet-18, and then choosing the best three structure with another five randomly formed structure to train and test on CIFAR-100. The results in Table \ref{tab: transfer of pss} show that the top 3 structures in CIFAR-10 also obtain the top 3 performances in CIFAR-100, which verifies the generalization of naturally formed parameter sharing structure. Compared to manually designed parameter sharing structure, our method is obviously more efficient.

\begin{table}[t]
\centering
\scriptsize
\setlength{\tabcolsep}{3.0pt}
\begin{tabular}{lcccccccc}
\hline
\multicolumn{9}{c}{CIFAR-10} \\ \hline
Architecure & 1 & 2 & 3 & 4 & 5 & 6 & 7 & 8 \\
Score & \bf 94.55 & 94.50 & 94.46 & 94.48 & 94.45 & \bf 94.53 & \bf 94.63 & 94.45 \\
Rank & \bf 2 & 4 & 6 & 5 & 7 & \bf 3 & \bf 1 & 7 \\ \hline
\multicolumn{9}{c}{CIFAR-100} \\ \hline
Architecure & 1 \bf(1) & 2 & 3 & 4 & 5 \bf(6) & 6 & 7 & 8 \bf(7) \\
Score & \bf 77.94 & 77.82 & 77.93 & 77.87 & \bf 77.97 & 77.81 & 77.88 & \bf 78.15 \\
Rank & \bf 3 & 7 & 4 & 6 & \bf 2 & 8 & 5 & \bf 1 \\ \hline
\end{tabular}
\caption{Transfer of parameter sharing structure. Architecture: ResNet-18. ``(\textbf{\#})'' is corresponding to the index of architectures on CIFAR-10.}
% \vspace{-2.0\baselineskip}
\label{tab: transfer of pss}
\end{table}

%%%%%%%%%%%%%%%%%%%%%%%%%%%%%%%%%%%%%%%%%%%%%%%%%%%%%%%%%%%%%%%%%%%%%%%%%%%%%%%%%%
\begin{figure}
 	\centering
	\includegraphics[width=0.45\linewidth]{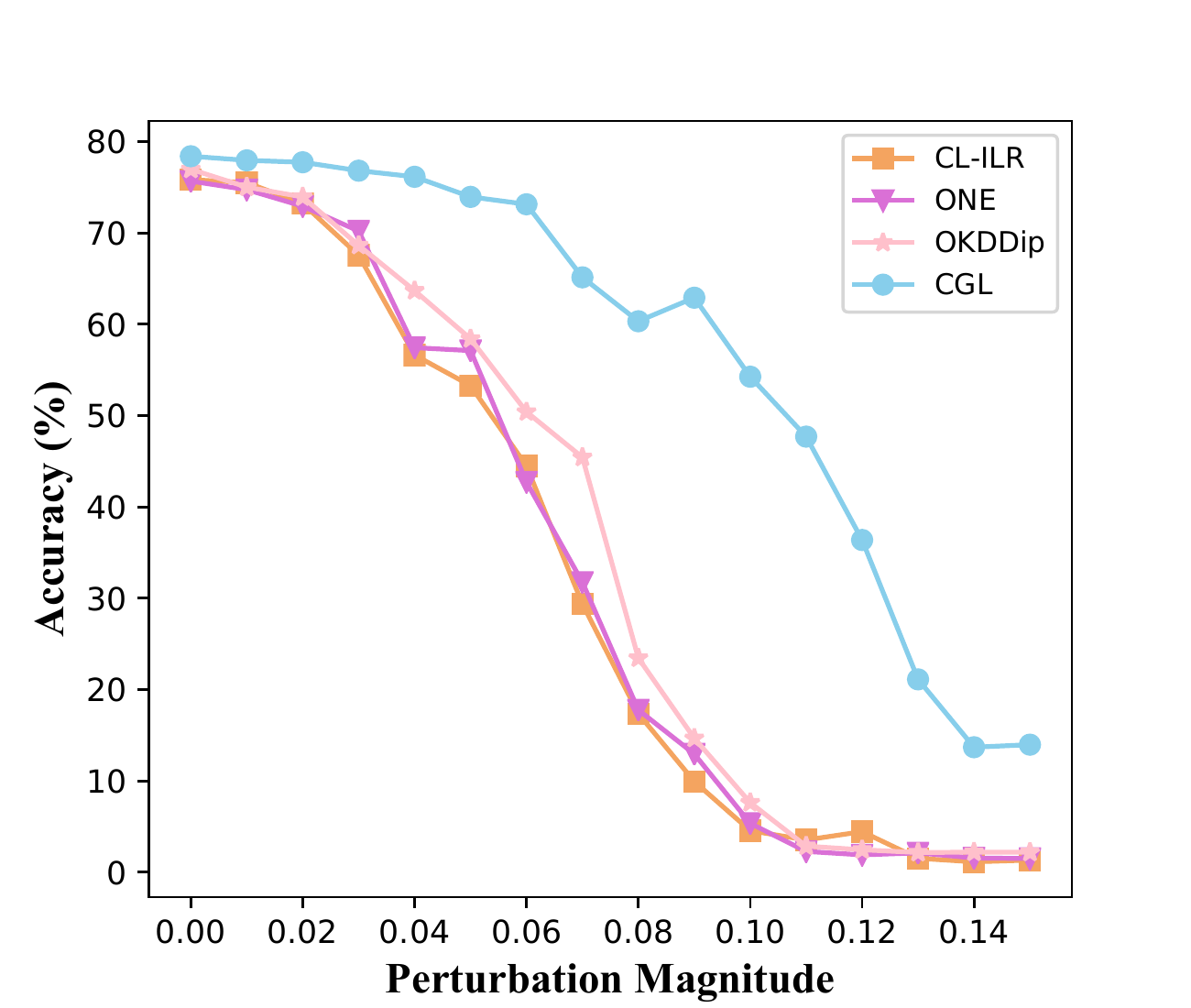} 
	\caption{Model generalization analysis.}
	% \vspace{-1.0\baselineskip}
	\label{fig: model generalization analysis}
\end{figure}
%%%%%%%%%%%%%%%%%%%%%%%%%%%%%%%%%%%%%%%%%%%%%%%%%%%%%%%%%%%%%%%%%%%%%%%%%%%%%%%%%%

\paragraph{Model Generalization Analysis}
We demonstrate why our collaborative group learning obtains better generalization than comparison methods. Recently, a collection of work \cite{DBLP:conf/iclr/ChaudhariCSLBBC17,DBLP:conf/iclr/KeskarMNST17} has proved that comparing a wider local minimum with a narrow one, the former is more beneficial for the model resisting small perturbations that dramatically damage the model accuracy. 
Inspired by this insight, we manually inject perturbations into the models to measure the width of local minima reached by all methods. Specially, we first generate different magnitudes of perturbations drawn from independent Gaussian distribution with variable standard deviation $\sigma$, and then add them to the model parameters. In Figure \ref{fig: model generalization analysis}, we plot the accuracy drop under different perturbation magnitudes. We can see that the accuracy of comparison methods declines much faster when the perturbation magnitude becomes larger. In contrast, our method is more stable, reflecting that collaborative group learning imposes effective regularization on the feature learning and guides the model towards a wider local minimum.
\section{Related Work}

\paragraph{Knowledge Distillation}
To deploy high-performance neural networks on mobile devices and embedded systems, Knowledge Distillation (KD) \cite{DBLP:conf/kdd/BucilaCN06,DBLP:conf/nips/BaC14,DBLP:journals/corr/HintonVD15,DBLP:journals/corr/RomeroBKCGB14} has been proposed to transfer fine-grained and hierarchical knowledge from a pre-trained large model (teacher) to a small model (student) by aligning the predictions or intermediate features of teacher and student. The student not only obtains similar performance as the teacher but also is easily deployed to the limited computation environment. Recently, several works \cite{DBLP:conf/iclr/ZagoruykoK17,DBLP:conf/cvpr/YimJBK17,DBLP:conf/icml/SrinivasF18,DBLP:conf/cvpr/AhnHDLD19} try to design new forms of teacher-learned knowledge or feature matching loss to facilitate knowledge transfer. KD suffers from pre-training a large teacher, which consumes more computational resources and training time; whereas we resort to collaborative learning and distill knowledge from a random sub-group of peer students.

\smallskip
\myparagraph{Collaborative Learning}
Collaborative learning \cite{DBLP:conf/cvpr/ZhangXHL18,DBLP:conf/nips/SongC18,DBLP:conf/nips/LanZG18,DBLP:conf/aaai/ChenMWF020} is more lightweight than KD in terms of learning stages. It facilitates each student to find a robust local minimum to achieve better generalization performance \cite{DBLP:conf/iclr/ChaudhariCSLBBC17,DBLP:conf/iclr/KeskarMNST17} in comparison to KD. Currently, there are two mainstream implementations of student networks. One is network-based \cite{DBLP:conf/cvpr/ZhangXHL18}, where students are independent networks, and the parameter capacity increases linearly with the number of students; the other is branch-based (CL-ILR \cite{DBLP:conf/nips/SongC18} and ONE \cite{DBLP:conf/nips/LanZG18}), where the bottom layers of students are shared. In our framework, we enable more flexible representation sharing with random routing mechanism \cite{DBLP:journals/corr/FernandoBBZHRPW17, DBLP:conf/nips/RajasegaranH0K019}, where layers at any level can be shared by different involved students. More importantly, students can be constructed as many as possible under restricted computational resources, whereas previous collaborative learning approach is more resource-intensive.

In terms of knowledge distillation in collaborative learning, OKDDip \cite{DBLP:conf/aaai/ChenMWF020} aggregates knowledge of all the students through weighted average. In contrast, we alleviate the student homogenization and enhance the model generalization ability by distilling knowledge from a random and dynamic sub-group of students, and each student learns different parts of the training data.
\section{Conclusion}
In this work, we present a novel knowledge distillation-based learning paradigm, collaborative group learning, which obtains better generalization performance and consumes lower computational cost than prior collaborative approaches. Specifically, adopting random routing to build students not only is more parameter-efficient but also enables flexible knowledge communication between students.
Besides, building more students promises more diverse knowledge at the beginning of training. To alleviate the student homogenization problem during training, sub-set data learning is introduced to diversify the feature sets of students, and sub-group imitation further boosts the diversity of group-level knowledge as well as enhances computational efficiency. Overall, our framework generates dynamic and diverse multi-view representations for the same input that effectively regularize the feature learning. Extensive experiments validate the effectiveness and robustness of our framework, and detailed analysis further proves that maintaining a balance between diversifying feature sets and internalizing group knowledge is essential for collaborative learning.

\section{Acknowledgements}
This research is supported by Beijing Natural Science Foundation (No. L181010 and  4172054), National Key R\&D Program of China (No. 2016YFB0801100), National Basic Research Program of China (No. 2013CB329605), and Beijing Academy of Artificial Intelligence (BAAI). Xu Sun and Kan Li are the corresponding authors.

\bibliography{aaai}
\bibstyle{aaai21}

\clearpage
\appendix
% \begin{appendices}

\section{Appendix}

\subsection{Computational Efficiency}

\paragraph{Analysis of Computational Cost} Take 8 students in each collaborative method and 16 batches of data in one epoch of training as an example. Previous collaborative methods consume $16*8*(forward, backward)$ steps for one epoch of training. 
Each student in CGL consumes $(N/K)*(1+K*p)*(forward, backward)$ steps, and its computational cost is $(2)*(1+2)*(forward, backward)$ steps when the imitation probability $p$ is 0.25. For one epoch of training, CGL can maintain constant $(forward, backward)$ steps by reducing the imitation probability $p$ when the number of students rises. Therefore, the increasing number of students will \textit{not affect the computational cost of our method} because of the adjustable imitation probability $p$ and the variable size $N/K$ of sub-set data.

\paragraph{Analysis of Training Time} Students in CGL can be \textit{run in parallel}, which means the number of students almost has no effect on the training time. Specifically, we trains $K+1$ copies of the modular neural network in parallel. Each student is corresponding to one selected module path in the assigned copy. In each iteration, we aggregate error gradients from the selected module paths of activated copies, update parameters of the $(K+1)^{th}$ copy, and then synchronize parameters of all other copies.

\paragraph{Analysis of Convergence Time} Besides, according to the statistics of raw experiment logs in Table 2 and Table 3 of the main paper, we find that there are \textit{no significant changes in the convergence speed} among all collaborative methods. The fluctuation range of convergence speed of CGL is $[-5.5\%, 6.5\%]$ compared to the average convergence speed of all baselines. 

% \end{appendices}

\end{document}